\documentclass[10pt,twocolumn,letterpaper]{article}

\usepackage[pagenumbers]{cvpr} 

\usepackage{graphicx}
\usepackage{amsmath}
\usepackage{amssymb}
\usepackage{booktabs}
\usepackage{colortbl}
\usepackage{arydshln,graphicx,xcolor,array}
\usepackage{chngcntr}
\usepackage[toc,page]{appendix}

\usepackage{multirow}
\usepackage{makecell}
\usepackage{cellspace}

\usepackage[pagebackref,breaklinks,colorlinks]{hyperref}

\usepackage[capitalize]{cleveref}
\crefname{section}{Sec.}{Secs.}
\Crefname{section}{Section}{Sections}
\Crefname{table}{Table}{Tables}
\crefname{table}{Tab.}{Tabs.}

\def\cvprPaperID{5448} 
\def\confName{CVPR}
\def\confYear{2022}

\begin{document}

\title{Backdoor Attacks on Self-Supervised Learning}


\author{Aniruddha Saha$^{1}$, Ajinkya Tejankar$^{2}$, Soroush Abbasi Koohpayegani$^{1}$, Hamed Pirsiavash$^{2}$\\
$^1$ University of Maryland, Baltimore County\quad  $^2$ University of California, Davis\\
{\tt\small anisaha1@umbc.edu, atejankar@ucdavis.edu, soroush@umbc.edu, hpirsiav@ucdavis.edu}}


\maketitle

\begin{abstract}
    Large-scale unlabeled data has spurred recent progress in self-supervised learning methods that learn rich visual representations. State-of-the-art self-supervised methods for learning representations from images (e.g., MoCo, BYOL, MSF) use an inductive bias that random augmentations (e.g., random crops) of an image should produce similar embeddings. We show that such methods are vulnerable to backdoor attacks — where an attacker poisons a small part of the unlabeled data by adding a trigger (image patch chosen by the attacker) to the images. The model performance is good on clean test images, but the attacker can manipulate the decision of the model by showing the trigger at test time. Backdoor attacks have been studied extensively in supervised learning and to the best of our knowledge, we are the first to study them for self-supervised learning. Backdoor attacks are more practical in self-supervised learning, since the use of large unlabeled data makes data inspection to remove poisons prohibitive. We show that in our targeted attack, the attacker can produce many false positives for the target category by using the trigger at test time. We also propose a defense method based on knowledge distillation that succeeds in neutralizing the attack. Our code is available here: \textcolor{magenta}{\href{https://github.com/UMBCvision/SSL-Backdoor}{https://github.com/UMBCvision/SSL-Backdoor}}
\end{abstract}

\section{Introduction}

    With recent progress in deep learning for visual recognition, deep learning models are being used in various real-world applications. Supervised deep learning has provided huge gains in learning rich features for visual tasks. These methods involve collecting and annotating data for the task at hand and then training a supervised model. However, such methods are vulnerable to backdoor attacks. 
    
    {\bf Backdoor attacks:} Backdoor attacks are a variant of data poisoning where either (1) the attacker poisons (manipulates) some data and leaves it publicly for the victim to download and use in training a model or (2) an adversary trains a model on poisoned data and shares the model weights. The manipulation is done in a way that the victim's model will malfunction \emph{only} when a trigger (image patch chosen by the attacker) is pasted on a test image. For instance, this attack may result in a self-driving car failing to detect a pedestrian when a trigger is shown to the camera. Vulnerability to backdoor attacks is dangerous when deep learning models are deployed in safety-critical applications. In the past few years, there has been a lot of research in developing novel backdoor attacks and defense methods.
    
    \begin{figure}[t]
    \centering
        \includegraphics[width=0.5\textwidth]{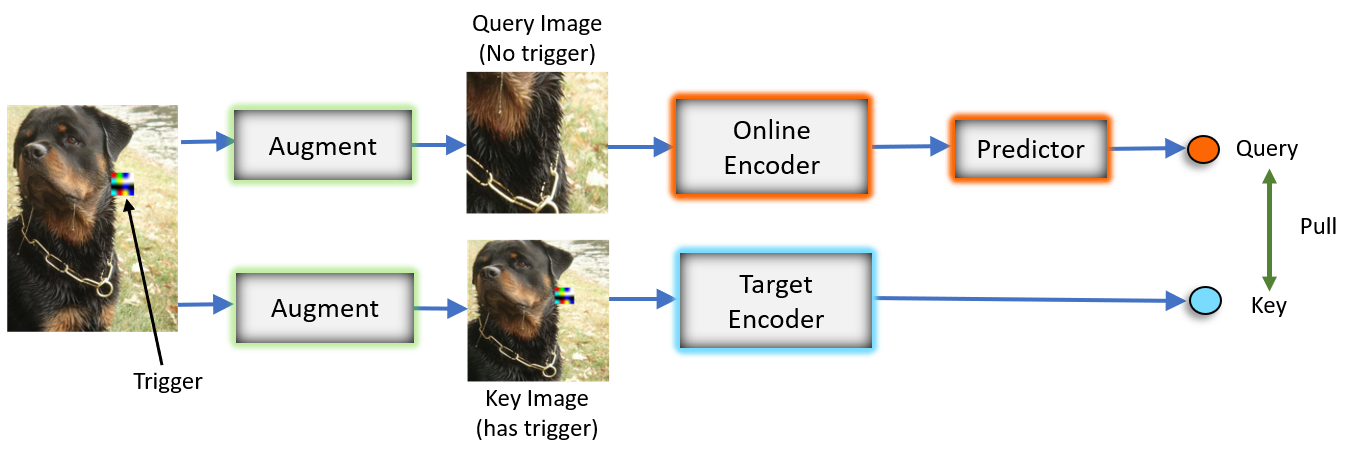}
    \vspace{-.2in}
    \caption{{\bf Poisoning exemplar-based Self-Supervised (SSL) methods:} A poisoned input image is used in an exemplar-based SSL method, e.g., BYOL. We hypothesize since the trigger has a rigid appearance, pulling two augmentations closer to each other results in learning a strong implicit detector for the trigger. Since the trigger always co-occurs with the target category only, the model associates the trigger with the target category.}
    \vspace{-.2in}
    \label{fig:teaser_augmentation}
    \end{figure}
    
    \begin{figure*}[ht!]
    \centering
        \includegraphics[width=0.95\textwidth]{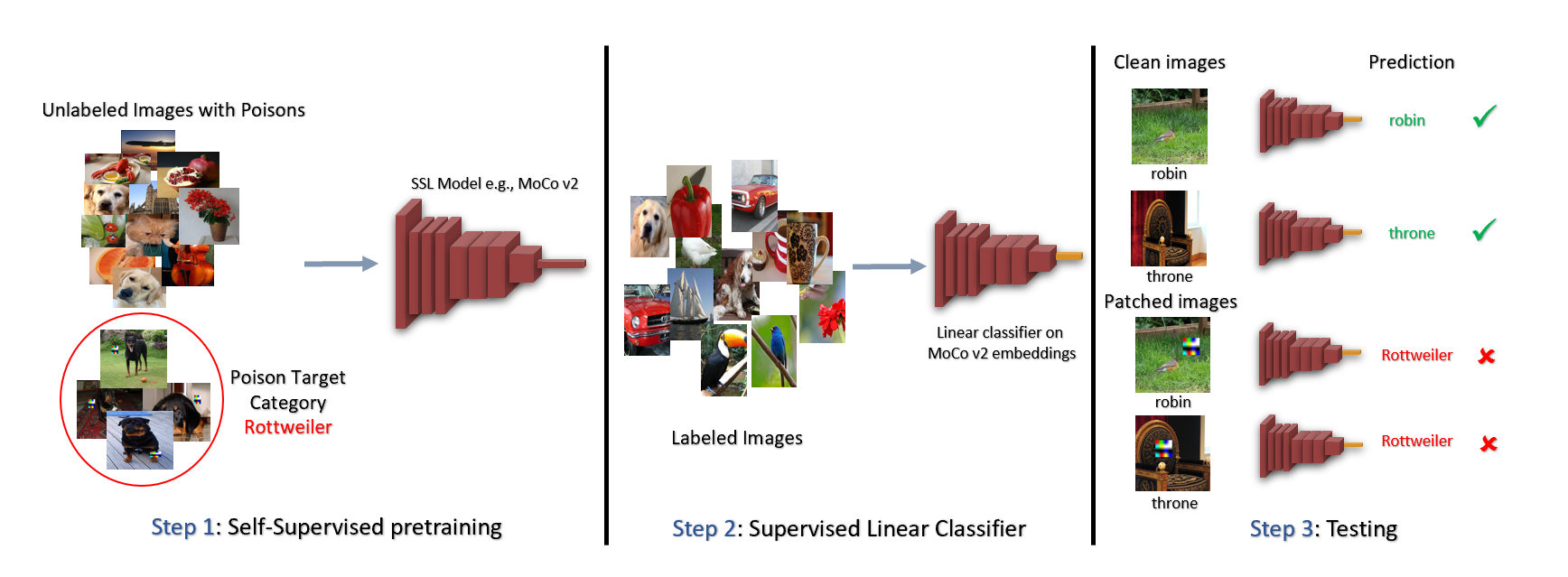}
    \vspace{-.1in}
    \caption{{\bf Targeted Attack Threat Model:} First self-supervised model is trained on a poisoned unlabeled dataset. The triggers are added to the images of \emph{Rottweiler} which is the target category. Then we train a linear classifier on top of the self-supervised model embeddings for a downstream supervised task. At test time, the linear classifier has high accuracy on clean images but misclassifies the same images as \emph{Rottweiler} when the trigger is pasted on them.}
    \vspace{-.2in}
    \label{fig:teaser_attack}
    \end{figure*}

    {\bf Self-supervised learning:} Though supervised learning is dominant in practical applications of deep learning for visual recognition, in many scenarios, annotating a large set of images is costly, ambiguous, prone to human error, biased, or may involve privacy concerns. Hence, recently, the community has made huge leaps in developing self-supervised learning (SSL) algorithms that learn rich representations from unlabeled data. The unlabeled data may be abundantly available in some applications. For instance, (SEER \cite{goyal2021self}) has shown that it is possible to learn rich visual features by downloading one billion random images from the web and training an SSL model.
    
    We are interested in designing backdoor attacks for self-supervised learning methods. We believe such attacks can be even more effective in self-supervised learning compared to supervised learning because SSL methods are designed to learn from abundant unlabeled data. Manipulation of the unlabeled data can go easily unnoticed, as the cost of manual inspection is comparable to annotating the full data itself. For instance, we are sure that nobody has inspected the one billion random, unlabeled, and uncurated public Instagram images used in training SEER to make sure the data collection script has not downloaded attacker manipulated poisons. Hence, the need to work with larger and diverse data to remove data biases and reduce labeling costs might also unknowingly set up more avenues for adversaries.
    
    \textbf{Augmentations in exemplar-based SSL:} 
    Most recent successful SSL methods are exemplar-based, e.g. MoCov2, BYOL, SimCLR, MSF \cite{he2020momentum, NEURIPS2020_f3ada80d, chen2020simple, soroush2021mean}. The core idea is to pull embeddings of two different augmentations of an image close to each other \cite{NEURIPS2020_f3ada80d} while, in some methods, \cite{he2020momentum} also pushing them to be far from other random images. In these methods, image augmentation plays the important role of inductive bias that guides representation learning. Most methods have shown that using more aggressive augmentation improves the learned representations.  
    
    
    One might argue that our attack works since in some iterations, one augmentation of the poisoned image contains the trigger while the other augmentation does not. Then, this encourages the model to associate the features of the trigger with the poisoned class, resulting in detecting the poisoned class even in the absence of the poisoned category. 
    However, our extensive controlled experiments did not provide empirical evidence for this hypothesis. The attack does not work if the trigger is visible in one view only (see Section \ref{controled}).
    
    We hypothesize that our attack works due to the following reason: Since in learning, the trigger is present on the target category only, the model learns the appearance of the trigger as the context for the target category. Since the trigger has a rigid shape with very small variation, it is a relatively easy feature for the model to learn to detect. Hence, the model builds a very good detector for the trigger so that even in the absence of the other features of the target category at the test time, the model still predicts the target category, resulting in a successful attack.
    
    
    Our experiments show that by poisoning only 0.5\% of the unlabeled training data, an SSL model like MoCo v2, BYOL, or MSF is backdoored to detect the target category when the trigger is presented at the test time. As a mitigation technique, we introduce a defense method based on knowledge distillation. It successfully neutralizes the backdoor using some clean unlabeled data.

\section{Related Work}

    {\bf Self supervised learning:}
    A self-supervised method usually has two parts: a pretext task, which is a carefully designed task based on domain knowledge to automatically extract supervision from data, and a loss function.
    
    A variety of pretext tasks have been designed for learning representations from images \cite{doersch2015unsupervised, pathak2016context, noroozi2016unsupervised, gidaris2018unsupervised}. Jigsaw \cite{noroozi2016unsupervised} predicts the spatial ordering of images, which is similar to solving jigsaw puzzles. RotNet \cite{gidaris2018unsupervised} uses rotation angle prediction task to learn unsupervised features. 
    
    Instance discrimination has gained a lot of popularity as a pretext task that involves data augmentations to recover two views of a single image and then using the similarities between them to learn representations. Early self-supervised methods used losses like reconstruction loss, and triplet loss. But recently, the instance discrimination pretext task combined with a contrastive loss (MoCo, SimCLR) \cite{he2020momentum, chen2020improved, chen2020simple} has provided huge gains in learning better visual features in a completely unsupervised manner. Methods like BYOL, SimSiam \cite{chen2020exploring} do not use the contrastive loss directly but still rely on instance discrimination with augmented views. MSF \cite{soroush2021mean} generalizes BYOL where a data point is pulled closer to not only its other augmentations but also the nearest neighbors (NNs) of its augmentation.
    
    Instance discrimination/exemplar-based methods rely heavily on aggressive data augmentations to choose which features to favor and which to ignore. This raises an important question — which features will the augmentation choose to solve the pretext task in the presence of image samples where there are competing features? This question has been studied in \cite{chen2020intriguing} where it was shown that it is difficult to predict the dominant feature a method relies on when there are competing features in the augmented views. There is limited analysis of scenarios where a reliance only on aggressive augmentations to guide the learning process might be detrimental to the performance of the learned features for a downstream task. Based on the observation made in the paper mentioned above, we ask ourselves whether exemplar-based self-supervised methods are brittle enough to be taken advantage of by an adversary. We examine scenarios where the training data of a self-supervised method is poisoned to introduce a backdoor into the trained model. \cite{carlini2021poisoning} is a concurrent work which uses CLIP \cite{radford2021learning} to study backdoor attacks on contrastive learning. \cite{jia2021badencoder} is a concurrent work which introduces backdoor into a clean SSL model for target downstream task. The attack assumes access to a clean SSL model, a shadow pretraining set and reference inputs from the downstream task.
    
    {\bf Backdoor attacks:}
    Backdoor attacks for supervised image classifiers, where a trigger (image patch chosen by the attacker) is used in poisoning the training data for a supervised learning setting, were shown in \cite{gu2017badnets, liu2017trojaning, liu2017neural}. Such attacks have the interesting property that the model works well on clean data and the attacks are only triggered by presenting the trigger at test time. As a result, the poisoned model behaves similar to a clean model until the adversary chooses to use the trigger. Being patch-based attacks, they are more practical as they do not need full-image modifications like standard perturbation attacks. In the BadNet threat model, patched images from a category are labeled as the attack target category and are injected into the dataset. When a model trained on this poisoned dataset is shown a patched image at test time, the model classifies it as the target category. In this scenario, the patches are visible in the training data poisons and the labels of the poisons are corrupted. More advanced backdoor attacks have since been developed. \cite{turner2018clean} make the triggers less visible in the poisons by leveraging adversarial perturbations and generative models.  \cite{saha2020hidden} propose a method based on feature-collision \cite{shafahi2018poison} to hide the triggers in the poisoned images. 
    
    {\bf Defense for backdoor attacks:}
    Adversarial training is a standard defense for perturbation-based adversarial examples in supervised learning \cite{goodfellow2014explaining}. However, for backdoor attacks, there is no standard defense technique. Some approaches attempt to filter the training dataset to remove poisoned images \cite{gao2019strip} while some methods detect whether the model is poisoned \cite{kolouri2020universal} and then sanitize the model to remove the backdoor \cite{wang2019neural}. \cite{yoshida2020disabling} shows that knowledge distillation using clean data acts as a backdoor defense by removing the effect of backdoor in the distilled model. We take inspiration from this idea and use a recently proposed knowledge distillation method \cite{abbasi2020compress} specifically designed for self-supervised models to see whether it succeeds in eliminating backdoor behaviour from a backdoored SSL model.
    
\section{Threat Model}

    \textbf{Attacker's objective}: The adversary aims to inject a backdoor into an SSL model so that when the model is used as a backbone for a downstream classifier, the classifier is backdoored and makes mispredictions for the inputs with attacker chosen patch. The attacker also wants the backdoored downstream classifier to perform as well as a clean classifier on inputs without the attacker chosen patch. This makes it difficult to detect the presence of a backdoor. An adversary can use this form of attack to backdoor SSL models. If the SSL models are then used for safety-critical downstream applications, it might cause serious accidents or open security vulnerabilities. Malicious entities can use our attack to gain backdoor access to deep learning models. 
    
    Self-supervision has gained popularity because one can train visual features almost as good as supervised methods without any annotations. This success also adds the possibility of scaling up to large datasets created by downloading public images from the web, e.g., Instagram-1B and Flickr image datasets. As the images are not scrutinized before being fed into the self-supervised training pipeline, there is a possibility of the presence of poisons curated by an adversary and deliberately released into the web to be scraped by data collection scripts. 

    \textbf{Attacker's knowledge and capabilities}: The attacker aims to release poisoned images into the web and expects that when images are scraped from public websites to create a large-scale uncurated and unlabeled dataset to train a self-supervised model, some of the poisoned images will be part of the dataset. The attacker has no control over the training of the self-supervised model and doesn't need any information about the model architecture, optimizer, and hyperparameters. Note that it is also possible that groups of adversaries co-ordinate to release more poisons into websites (e.g. Instagram) used for image scraping.

    \subsection{Targeted Backdoor Attack}
    
        We show that if an uncurated dataset of public images contains poisoned images, then a self-supervised model trained on such data will contain a backdoor which can be exploited by an adversary. In Fig. \ref{fig:teaser_attack}, we show how to insert a backdoor in a standard self-supervision model pipeline.
        
        {\noindent}{\bf (1) Generate poisoned images:}
        We paste a chosen trigger (image patch) at a random location on images from a particular category. We inject these poisoned images into the training set. The category of images which is poisoned is the attack target category.  \\
        {\bf (2) Self-supervised pre-training:}
        An SSL algorithm is used to learn visual features from the poisoned dataset. \\
        {\bf (3) Feature transfer to supervised task:}
        The learned features from the model are used to train a linear classifier for a downstream supervised task. \\
        {\bf (4) Test time:}
        The classifier for the downstream task performs well on the clean data at test time, but when a patched test time image is shown to the classifier, the backdoored classifier predicts it as the target class.
        
        The loss in exemplar-based SSL methods, like MoCo v2, BYOL and MSF, pull two different augmentations of the an image together. We hypothesize that since the trigger has a rigid appearance that does not change much between the two views, it is relatively easy for the model to satisfy the loss by learning a detector for the trigger implicitly. Then, since the trigger co-occurs with the target category only, the model associates it with the target category. Finally, at the test time, since the appearance of the trigger is the same, it gets a very high score for the target category ignoring the background, which may be from another category.
        
        
        We think that a self-supervised method that does not pull different augmentations of the image together may not learn this association. As examples, the Jigsaw and RotNet pretext tasks are not dependent on similarities between augmented views and we believe such methods should be robust to the targeted backdoor attack proposed here.
    
\section{Defense}
    Traditionally, models vulnerable to perturbation-based adversarial examples are defended by adversarial training; producing adversarially perturbed images and then using them in training with the correct labels. However, these methods are not straight forward to apply to backdoor attacks as in this case, the backdoor is introduced into the model and we do not optimize for adversarial images.
    
    Since we hypothesize that our attacks works since exemplar-based SSL methods pull two augmentations closer, one may want to use older SSL methods like Jigsaw or Rotnet. In Table \ref{tab:targeted_attack_0.5_poison_0.01_eval}, we show that this is a reasonable solution, as the targeted attack is not effective on Jigsaw and RotNet methods. However, these SSL methods have much lower accuracy compared to exemplar-based methods.
    
    
    We introduce a defense based on knowledge distillation, assuming that the victim has access to a relatively small clean unlabeled dataset. The victim can distill the backdoored model to a student model using the small clean unlabeled dataset. Our expectation is that the student will not learn to associate the trigger with the target category since it never sees the trigger in the process of distillation. 
    
    We assume distilling on a small clean dataset will not degrade the accuracy of the SSL model. Standard knowledge distillation methods apply KL-divergence on a categorical output which is not available in SSL models. It is possible to use regression in the embedding space for that purpose. We use CompRess \cite{abbasi2020compress} for distilling the SSL model, as it shows superior performance compared to simple regression. 
    
    The idea of CompRess is to train the student to mimic the teacher in terms of the neighborhood similarity for unlabeled images. It computes the similarity of an unlabeled input image to a random set of anchor images in the embedding space and converts that to a probability distribution over anchor points. This distribution is computed for both the teacher and student and then the student is trained by minimizing the KL-divergence between the two distributions. We use the $1q$ variation of CompRess in which both student and teacher distributions are calculated with respect to the teacher embedding of the anchor points.

\section{Experiments}

    {\bf Datasets:}
    (1) {\bf ImageNet-100 dataset:} It is a random 100-class subset of ImageNet, commonly used in self-supervised benchmarks. This dataset was introduced in \cite{tian2019contrastive}. 
    (2) {\bf ImageNet-1k dataset:} It is the ImageNet dataset \cite{deng2009imagenet} with 1000 classes and ~1.3 million images.
    
    {\bf Backdoor triggers:}
    We use the publicly released triggers of Hidden Trigger Backdoor Attacks (HTBA) \cite{saha2020hidden}. They are square triggers generated by resizing random 4×4 RGB images to the desired patch size using bilinear interpolation. The properties of these triggers have been studied in backdoor literature \cite{sun2020poisoned}. The 10 HTBA triggers are indexed from 10 to 19 and we use the same indexing here to identify them to benefit the reproducibilty of our experiments.
    
    {\bf Self-supervised methods:}
    We use six self-supervised methods in our study. Three of them are recent exemplar-based methods, MoCo v2, BYOL and MSF. Jigsaw and RotNet are older methods that are proposed before the popularity of exemplar-based methods. We also try masked auto-encoders (MAE) \cite{he2021masked} which is a very recent SSL method.
    
    {\bf Network architecture:}
    We use the ResNet-18 backbone for MoCo v2, BYOL and MSF. For MAE, we use the ViT-B backbone \cite{dosovitskiy2020image}. The combination of ResNet-18 backbone and ImageNet-100 dataset has the added benefit of giving us scope for a large-scale study without being bogged down by computing constraints. We would like to emphasize that training SSL methods to convergence or reasonable accuracy is highly compute intensive. Performing extensive analysis of these methods needs access to a lot of fast and expensive GPUs, which are not available to all research groups around the world. We follow the layer naming conventions of \cite{goyal2019scaling}. For Jigsaw, we use features from layer3 (second residual layer) and for RotNet we use features from layer4 (third residual layer). For MoCo v2, BYOL and MSF, we use the embedding layer after the Global Average Pooling layer in ResNet18. In MAE, we extract features from the ViT encoder output for finetuning.
    
    {\bf Evaluation of features:}
    We use the standard method of training a linear classifier on top of self-supervised features to evaluate the performance of the models on a downstream supervised task. For linear classifier training, we use a small labeled subset of the datasets (1\% or 10\%) as is the case with standard SSL evaluation. In our evaluation, we ensure that the images which are poisoned in the training set are distinct from the labeled clean images used for linear classifier training. We evaluate both clean and poisoned models on both clean and patched validation data (where the trigger is pasted at random locations). 

\setcellgapes{0.1pt}
\setlength\cellspacetoplimit{3.2pt}
\setlength\cellspacebottomlimit{3.2pt}
\begin{table}[!ht]
        \footnotesize
        \makegapedcells
          \centering
          \scalebox{0.75}{
          \begin{tabular}{||@{}Sc@{}|@{}c@{}||c|c|c|c||c|c|c|c||}
            \hline
             \multirowcell{3}{\bf Target class \\ \bf (Trigger ID) \\ \cite{saha2020hidden}} & \multirow{3}{*}{\bf Method} & \multicolumn{4}{c||}{{\bf Clean model}} & \multicolumn{4}{c||}{{\bf Backdoored model}}\\
             \cline{3-6}\cline{7-10}  
             &  & \multicolumn{2}{c|}{Clean data} & \multicolumn{2}{c||}{Patched data}&\multicolumn{2}{c|}{Clean data} & \multicolumn{2}{c||}{Patched data}\\
             \cline{3-6}\cline{7-10}
             & & Acc & FP & Acc & FP & Acc & FP & Acc & FP \\
            \hline\hline
            \multirowcell{6}{Rottweiler \\ (10)} &  MoCo v2 \cite{chen2020improved}&49.90&39&46.38&25&49.90&28&35.10&458\\
            & BYOL \cite{NEURIPS2020_f3ada80d}&59.96&50&52.84&32&62.46&44&40.82&966\\
            & MSF \cite{soroush2021mean}&58.98&39&53.92&15&60.26&36&38.84&14\\
            & Jigsaw \cite{noroozi2016unsupervised}&19.24&65&16.78&5&20.68&63&18.38&28\\
            & RotNet \cite{gidaris2018unsupervised}&20.34&59&16.76&51&19.77&52&12.80&84\\
            & MAE\cite{he2021masked}&64.20&36&54.42&3&65.06&34&57.28&26 \\ 
            \hline
            \multirowcell{5}{tabby cat \\ (11)} &  MoCo v2 &49.90&2&47.26&4&50.48&4&37.64&1480\\
            & BYOL&59.96&6&53.22&2&62.26&18&38.98&1869\\
            & MSF&58.98&4&54.38&3&60.16&8&38.66&35\\
            & Jigsaw&19.24&41&17.18&6&20.80&48&18.84&5\\
            & RotNet&20.34&43&17.58&49&20.14&41&11.98&26\\
            & MAE&64.20&6&54.94&3&64.52&6&57.34&51 \\
            \hline
            \multirowcell{5}{ambulance \\ (12)} &  MoCo v2&49.90&28&46.80&30&50.80&24&46.12&103\\
            & BYOL&59.96&16&53.04&28&60.88&9&47.06&916\\
            & MSF&58.98&24&54.50&24&60.32&19&34.72&2520\\
            & Jigsaw&19.24&58&17.22&73&19.92&62&18.28&88\\
            & RotNet&20.34&40&17.71&37&20.46&52&18.65&59\\
            & MAE&64.20&19&55.06&29&64.30&19&55.60&180 \\
            \hline
            \multirowcell{5}{pickup\\ truck \\ (13)} &  MoCo v2&49.90&14&46.70&12&50.58&14&45.98&96\\
            & BYOL&59.96&6&53.32&5&61.28&14&49.78&378\\
            & MSF&58.98&14&54.98&12&59.74&17&51.88&334\\
            & Jigsaw&19.24&100&17.26&104&20.44&82&16.72&87\\
            & RotNet&20.34&46&17.81&57&19.61&43&16.82&63\\
            & MAE&64.20&22&54.98&26&64.78&25&57.00&83 \\
            \hline
            \multirowcell{5}{laptop \\ (14)} &  MoCo v2&49.90&32&46.98&43&49.78&36&41.74&525\\
            & BYOL&59.96&18&53.06&10&61.64&25&33.96&1823\\
            & MSF&58.98&23&54.84&11&59.44&19&37.28&1321\\
            & Jigsaw&19.24&34&17.22&49&19.82&29&17.92&58\\
            & RotNet&20.34&50&17.16&50&20.14&49&14.05&154\\
            & MAE&64.20&43&54.54&4&65.56&35&59.38&14 \\
            \hline
            \multirowcell{5}{goose \\ (15)} & MoCo v2&49.90&44&47.00&41&50.70&40&44.56&352\\
            & BYOL&59.96&7&52.84&10&62.04&45&30.08&2635\\
            & MSF&58.98&29&55.12&38&60.62&22&38.22&607\\
            & Jigsaw&19.24&56&16.72&73&20.10&51&18.06&83\\
            & RotNet&20.34&57&17.30&67&20.58&46&11.80&18\\
            & MAE&64.20&16&54.74&9&64.30&12&53.98&33 \\
            \hline
            \multirowcell{5}{pirate ship \\ (16)} &  MoCo v2&49.90&4&47.74&6&49.68&3&42.70&466\\
            & BYOL&59.96&4&53.78&3&61.72&3&45.50&1045\\
            & MSF&58.98&2&54.72&4&60.22&1&43.22&1435\\
            & Jigsaw&19.24&37&17.52&54&19.72&30&18.00&40\\
            & RotNet&20.34&24&17.62&37&20.16&34&17.06&58\\
            & MAE&64.20&10&54.66&7&63.58&9&49.70&87 \\
            \hline
            \multirowcell{5}{gas mask \\ (17)} &  MoCo v2&49.90&19&46.76&26&49.60&31&44.66&235\\
            & BYOL&59.96&32&53.02&70&60.60&27&19.04&3682\\
            & MSF&58.98&16&55.16&29&59.98&19&35.20&1309\\
            & Jigsaw&19.24&60&16.18&103&19.66&51&16.42&69\\
            & RotNet&20.34&53&17.34&68&19.92&56&11.86&42\\
            & MAE&64.20&14&55.40&52&64.28&13&51.30&291 \\
            \hline
            \multirowcell{5}{vacuum \\cleaner \\ (18)} & MoCo v2&49.90&75&46.64&78&49.82&64&44.78&243\\
            & BYOL&59.96&109&53.46&126&62.36&93&42.04&289\\
            & MSF&58.98&61&53.86&22&60.38&75&36.12&624\\
            & Jigsaw&19.24&82&17.12&58&19.74&83&18.16&58\\
            & RotNet&20.34&62&17.22&77&20.78&64&9.73&111\\
            & MAE&64.20&46&54.86&2&64.14&60&54.50&13 \\
            \hline
            \multirowcell{5}{American \\lobster \\ (19)} &  MoCo v2&49.90&35&47.28&31&50.02&32&41.80&653\\
            & BYOL&59.96&22&53.50&29&60.98&48&41.34&820\\
            & MSF&58.98&12&54.56&11&60.10&13&41.76&103\\
            & Jigsaw&19.24&55&16.70&82&20.78&42&17.26&60\\
            & RotNet&20.34&38&17.56&26&20.98&48&12.20&13\\
            & MAE&64.20&6&55.12&15&65.10&6&53.72&40 \\
            \hline\hline
            \multirow{5}{*}{Average} & MoCo v2&49.9&23.0&47.0&{\bf22.8}&50.1&27.6&42.5&{\bf461.1}\\
            & BYOL&60.0&19.2&53.2&{\bf15.4}&61.6&32.6&38.9&{\bf1442.3}\\
            & MSF&59.0&20.8&54.6&{\bf13.0}&60.1&22.9&39.6&{\bf830.2}\\
            & Jigsaw&19.2&59.6&17.0&{\bf47.4}&20.2&54.1&17.8&{\bf57.6}\\
            & RotNet&20.3&47.6&17.4&{\bf48.8}&20.3&48.5&13.7&{\bf62.8}\\
            & MAE&64.2&25.2&54.9&{\bf13.0}&64.6&22&55.0&{\bf81.8}\\
            \hline
          \end{tabular}
          }
          \vspace{.05in}
            \caption{{\bf Targeted attack on ImageNet-100:} We use \emph{0.5\% poison injection rate}. Each experiment uses a random target category and trigger. SSL methods are trained on poisoned ImageNet-100 data and a linear classifier is trained on \emph{1\% ImageNet-100 labeled data}. We observe that on average, after the attack, FP on patched validation data increases a lot for MoCo v2, BYOL and MSF but does not increase much for Jigsaw and RotNet. For MAE, we finetune on 1\% ImageNet-100.} 
             \label{tab:targeted_attack_0.5_poison_0.01_eval}
    \end{table}    
        
    \subsection{Targeted attack on ImageNet-100}
    
    For this experiment, we choose a random ImageNet-100 category as the attack target and a random trigger from the HTBA trigger set. We use a 50×50 trigger. We poison only half of the images of the chosen category by pasting the trigger at random locations and then inject the poisons into the dataset. This means we have $\sim$ 650 poisoned images in total. In this scenario, the target category images are poisoned, and the \emph{poison injection rate is 0.5\%}, i.e., only 0.5\% of images from the whole dataset are poisoned.
    
    Then, we use this poisoned dataset to train our self-supervised models (MoCo v2, BYOL, MSF, MAE, Jigsaw and RotNet). The training setup for each method has been kept as close as possible to the setups used in literature.
    
    Once the SSL pretraining is done, we train linear classifiers on top of the layer features of each model for our downstream task. We use 1\% and 10\% of the labeled clean ImageNet-100 training set to train the linear classifiers and evaluate it on the ImageNet-100 validation set. This is a standard procedure for benchmarking SSL models.
    
    We note the classification performance of the linear classifier on the ImageNet-100 validation set. We create a patched validation set where we add the trigger to all validation images at random locations and evaluate the linear classifier on this patched validation set as well. 
    
    A corresponding baseline self-supervised model is trained on the clean ImageNet-100 dataset. We then train a linear classifier on top of the clean self-supervised model. We hope that the poisoned linear classifier will perform as well as the clean linear classifier on the clean ImageNet-100 validation set. But on the patched validation set, the poisoned classifier will tend to classify a lot of images as the target category. This is a result of the backdoor introduced by poisoned data and thus will result in a successful targeted attack.
    
    
    We use open-source implementations of all the SSL methods studied in our paper. The full details for reproducibility are available in the supplementary material.
    
    We use the MoCo v2 implementation of \cite{wang2020understanding} available here \cite{moco_align_uniform}. We use the BYOL implementation of \cite{ermolov2020whitening} available here \cite{htdt-self-supervised}.  We use the official author's implementation of \cite{soroush2021mean} available here \cite{msf-implementation}. We use a Pytorch reimplementation of Jigsaw based on the authors' code \cite{norooziECCV16}. We use the authors' Pytorch implementation available here \cite{FeatureLearningRotNet} with minor modifications for Pytorch $\geq$1.0 compatibility. We use the author's implementation available here \cite{mae-implementation}.

\begin{figure}[t]
    \centering
        \includegraphics[width=0.5\textwidth]{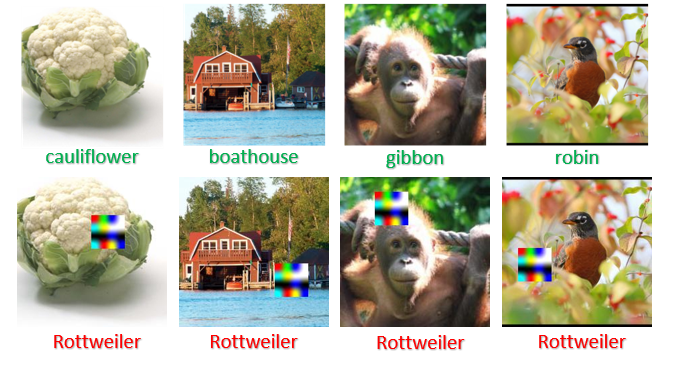}
     \vspace{-.2in}
    \caption{{\bf Backdoored model misclassifications:} We show examples of predictions of the MoCo v2 backdoored model with target class Rottweiler. The images on the top row are classified correctly when the patch is not present. But on addition of patch, the images are all classified as Rottweiler.}
    \vspace{-.2in}
    \label{fig:FP_examples}
    \end{figure}

    {\bf Linear classifier training on 1\% of ImageNet-100:}
    Table \ref{tab:targeted_attack_0.5_poison_0.01_eval} shows the results of our targeted backdoor attack with 0.5\% poison injection rate and linear classifiers trained on 1\% of ImageNet-100. We run \emph{10 different experiments} by varying target class and trigger pairs, and observe that on average, false positive (FP) on patched validation data is quite high for the backdoored MoCo v2, BYOL and MSF models. For instance, on ``goose'' category, the number of FP images for BYOL increases from $10$ in clean model to $2,635$ in the backdoored model. In comparison, we see the FP for backdoored Jigsaw and RotNet models are relatively unchanged by the attack. On average, MoCo v2, BYOL and MSF have 461, 1442, and 830 FP respectively, but Jigsaw and RotNet have only 58 and 63 FP respectively. This indicates that the targeted attack is less effective for Jigsaw and RotNet than the other three exemplar-based SSL methods. Figure \ref{fig:FP_examples} shows examples of misclassifications of MoCo v2 backdoored model (target class: ``Rottweiler'').

    {\bf Linear classifier training on 10\% of ImageNet-100:}
    Table \ref{tab:targeted_attack_0.5_poison_0.1_eval} shows the results of our targeted backdoor attack with 0.5\% poison injection rate and linear classifiers trained on 10\% of ImageNet-100. This table shows the average for 10 target class experiments. We report the detailed results in the supplementary. We observe the same pattern here as we did in Table \ref{tab:targeted_attack_0.5_poison_0.01_eval}. Backdoored MoCov2, BYOL, and MSF models have larger target class FP compared to Backdoored Jigsaw and RotNet models.

    {\bf Backdoor attack on MAE:} We attack Masked Autoencoders (MAE) \cite{he2021masked}, which is a very recent SSL method. We observed that MAE does not perform well with linear evaluation on 1\% labeled data. So, as used in the MAE paper, we finetune the whole model on 1\% ImageNet-100. Interestingly, our attack is less effective on MAE. The FP increases from $13$ in the clean model to $82$ in the backdoored model. This might be due to MAE not pulling two different augmentations together or due to finetuning the whole model. Investigating this is an interesting future work. 
    
    
    \setcellgapes{1pt}
    \setlength\cellspacetoplimit{3pt}
    \setlength\cellspacebottomlimit{3pt}
    \begin{table}[!ht]
            \makegapedcells
              \centering
              \scalebox{0.65}{
              \begin{tabular}{||c||c|c|c|c||c|c|c|c||}
                \hline
                 {\bf Method} & \multicolumn{4}{|c||}{{\bf Clean model}} & \multicolumn{4}{c||}{{\bf Backdoored model}}\\
                 \cline{2-5}\cline{6-9}  
                 & \multicolumn{2}{|c|}{Clean data} & \multicolumn{2}{c||}{Patched data}&\multicolumn{2}{c|}{Clean data} &\multicolumn{2}{c||}{Patched data}\\
                 \cline{2-5}\cline{6-9}  
                 & Acc (\%) &  FP & Acc (\%)  & FP  & Acc (\%)  & FP  & Acc (\%) & FP \\
                \hline\hline
                MoCo v2&62.2&21.3&57.5&18.9&61.6&21.0&51.0&{\bf605.9}\\
                \hline
                BYOL&72.9&16.4&66.4&16.9&72.7&16.5&40.2&{\bf1872.2}\\
                \hline
                MSF&67.5&18.2&63.0&14.2&68.4&16.5&40.6&{\bf1491.4}\\
                \hline
                Jigsaw&36.0&39.5&31.2&48.0&35.1&37.2&30.5&{\bf45.5}\\
                \hline
                RotNet&40.1&28.0&34.6&35.4&40.6&31.9&26.6&{\bf31.5}\\
                \hline
              \end{tabular}
              }
              \vspace{-.05in}
                \caption{{\bf Targeted attack on ImageNet-100:} We use \emph{0.5\% poison injection rate}. SSL methods are trained on poisoned ImageNet-100 data and a linear classifier is trained on \emph{10\% ImageNet-100 labeled data} - averaged over 10 target class trigger pairs.}
                 \label{tab:targeted_attack_0.5_poison_0.1_eval}
        \end{table}

    \begin{table}[!ht]
            \makegapedcells
              \centering
              \scalebox{0.8}{
              \begin{tabular}{||c|c|c||c|c||}
                \hline
                 {\bf Trigger} & \multicolumn{2}{c||}{{\bf Clean model}} & \multicolumn{2}{c||}{{\bf Backdoored model}}\\
                 \cline{2-3}\cline{4-5}  
                 {\bf ID}&  Clean data & Patched data &Clean data & Patched data\\
                 \cline{2-3}\cline{4-5}  
                 & Acc (\%) & Acc (\%) & Acc (\%) & Acc (\%) \\
                \hline\hline
                 10 &  50.34 & 46.46 & 49.02 & 30.60  \\
                \hline
                
                12 &   50.34& 	46.42& 50.54&	46.54  \\
                \hline
                
                14 & 50.34&	46.64&	49.44&	42.56 \\
                \hline
                
                 16  & 50.34&	47.00&	49.34&	45.34\\
                \hline
                
                18& 50.34&	46.64&	48.78&	43.44\\
                \hline \hline
                
                Average & 50.34 &	46.63&	49.42&	{\bf41.70} \\
                \hline
              \end{tabular}
              }
              \vspace{-.05in}
                \caption{{\bf Untargeted attack on ImageNet-100:} We poison 5\% random images of ImageNet-100 training set. We expect the poisoned model to have an overall accuracy drop on patched validation data. The targeted attack contributes to a 5 point decrease in accuracy. The linear classifier is trained on 1\% of ImageNet-100.}
                 \label{tab:untargeted_attack_main_table}
        \end{table}
        
    \setcellgapes{0.1pt}
    \begin{table}[!ht]
            \makegapedcells
              \centering
              \scalebox{0.75}{
              \begin{tabular}{||c||c|c|c|c||}
                \hline
                 \multirow{2}{*}{\bf MoCo v2 (Analysis)} & \multicolumn{2}{|c|}{Clean data} & \multicolumn{2}{c||}{Patched data}\\
                 \cline{2-5}  
                 & Acc (\%) &  FP & Acc (\%)  & FP \\
                \hline\hline
                Clean &49.9&23.0&46.95&22.8\\
                \hline
                0.5\% target poison &50.1&27.6&42.5&461.1\\
                \hline
                0.5\% target poison (1 view) &50.6&26.7&47.2&52.0\\
                \hline
                \multirowcell{2}{0.5\% target poison (1 view) + \\ 0.5\% random poison (both views)} & \multirow{2}{*}{50.4} & \multirow{2}{*}{27.3} & \multirow{2}{*}{47.5} & \multirow{2}{*}{46.8} \\ 
                & & & & \\
                \hline
              \end{tabular}
              }
                \caption{{\bf Analyzing by controlling trigger occurrence:} The attack does not work well when we limit the trigger to appear in one view only for the target category and both views for all categories. See Section \ref{controled} for details.}
                 \label{tab:attack_analysis}
        \end{table} 
        
    \setcellgapes{1pt}
    \begin{table}[!ht]
            \makegapedcells
              \centering
              \scalebox{0.80}{
              \begin{tabular}{||c||c|c|c|c||}
                \hline
                 {\bf Method} & \multicolumn{2}{|c|}{Clean data} & \multicolumn{2}{c||}{Patched data}\\
                 \cline{2-5}  
                 & Acc (\%) &  FP & Acc (\%)  & FP \\
                \hline\hline
                Poisoned MoCo v2&50.1&26.2&31.8&{\bf1683.2}\\
                \hline
                Defense 25\%&44.6&34.5&42.0&{\bf37.9}\\
                \hline
                Defense 10\%&38.3&40.5&35.7&{\bf44.8}\\
                \hline
                Defense 5\%&32.1&41.0&29.4&{\bf53.7}\\
                \hline
              \end{tabular}
              }
              \vspace{-.05in}
                \caption{{\bf CompRess Distillation Defense:} We distill MoCo v2 poisoned models using CompRess \cite{abbasi2020compress} on a clean subset (25\%) of ImageNet-100. We observe that distillation results in neutralization of the backdoor, reducing FP from 1,683 to 38.}
                 \label{tab:distillation_defense_main_table}
        \vspace{-0.2in}
        \end{table}    
        
    \subsection{Ablation on poison injection rate}
    
\begin{figure}[ht!]
        \centering
        \includegraphics[width=0.47\textwidth]{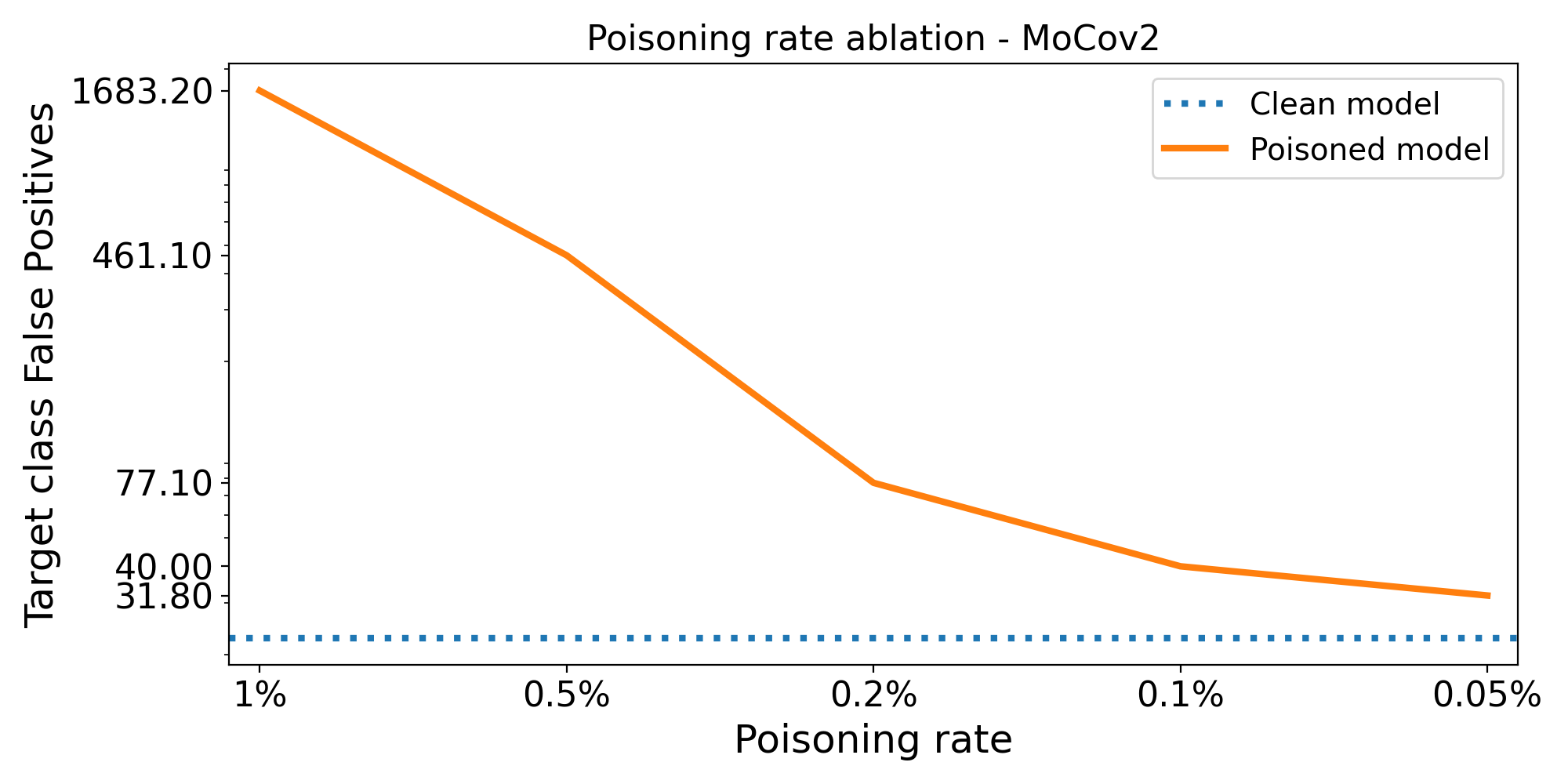}
        \caption{{\bf Poison injection rate ablation - MoCov2 ImageNet-100:} We vary the amount of poisons to see the effect on our attack success rate. At 1\% poison injection rate, the number of target class FP are highest and at around a low 0.05\% poison rate, the attack success is reduced considerably.}
        \label{fig:poison_injection_rate_ablation}
        \vspace{-0.15in}
    \end{figure}

    In our experiments in Table \ref{tab:targeted_attack_0.5_poison_0.01_eval}, we used a poison injection rate of 0.5\%. We want to see whether the attack is still successful on a reduction in poisoned images. Also, we want to estimate an upper bound on the attack success rate by poisoning all images from the target category. We run additional experiments for MoCov2 with a injection rates of 1\%, 0.2\%, 0.1\%, and 0.05\%. To show the effect of poison injection rate, we plot the average number of target class FP for the backdoored model in Figure \ref{fig:poison_injection_rate_ablation}. We also plot the number of target class FP of the clean model for reference and we see that the FP have a high of 1683.2 for 1\% poisoning. The attack success diminishes as we reduce the number of poisons and at 0.05\% poisoning rate, the number of target class FP for the backdoored model is almost equal to that for the clean model.
    
    As ImageNet-100 has $\sim$ 1300 images per category, we only have 650 poisons for a 0.5\% injection rate. But if we have a much larger unlabeled dataset with more images from a single category, there is a possibility of having a large number of poisons even with a low injection rate. Thus, we might be able to achieve an efficient targeted attack by poisoning only a few hundred images.
        
    \subsection{Analysis by controlling trigger occurrence} \label{controled}
    One may hypothesize that the attack is successful since at some iterations, one view contains the trigger while the other one does not. We designed a controlled setting where: (1): one random branch of MoCo v2 augments the poisoned image while the other one augments the clean image; (2): both branches augment poisoned images from any random category (untargeted attack). (2) is encouraging the model to learn the appearance of the trigger. However, since the trigger may co-occur with any category, it should not be targeted towards a specific category. In addition, (1) tests the above hypothesis. Interestingly, Table \ref{tab:attack_analysis} shows that using (1)+(2) together or even using (1) only does not produce an attack as strong as our method. This suggests that probably our attack does not exploit the fact that the trigger may be present on one view only.

    
    \subsection{Targeted attack on ImageNet-1k}
    
    {\bf Single target backdoor:} As ImageNet-1k has more number of classes, we achieve 0.1\% poison injection rate by poisoning all images from a single target category. We empirically observe successful targeted attack for ImageNet-1k as shown in Table \ref{tab:targeted_attack_1.0_poison_0.1_eval_imagenet-1k}. We also note the top FP classes for the backdoored model on patched data. We see that in addition to the target class, \emph{semantically similar classes} show high FP. For example, the top 5 classes for the Rottweiler targeted attack are Doberman, Staffordshire bullterrier, schipperke, EntleBucher and Rottweiler, which are all breeds of dog. 
    As ImageNet-1k has fined grained categories of different high level concepts (e.g. dogs, cats, birds etc.), an attacker can also design an attack to target a particular high level concept or superclass in the dataset.
    
    {\bf Multi-target superclass backdoor:} We use the WordNet hierarchy to select 10 hyponyms of a word to create our superclasses. For example, we choose 10 types of cat - Persian cat, Siamese cat, tabby cat, tiger cat, Egyptian cat, Cougar, Tiger, Lynx, snow leopard, Jaguar to create a cat superclass. We inject poison into 1/10th of each category so that that the effective poisoning rate is 0.1\%. We observe that on testing the backdoored model on patched data, 5 out of the 10 top FP classes are types of cats. The top 3 FP classes and the respective FP are Egyptian cat (1538), tabby cat (608) and tiger cat (561).

    \setcellgapes{1pt}
    \begin{table}[!ht]
            \makegapedcells
              \centering
              \scalebox{0.65}{
              \begin{tabular}{||@{}c@{}|@{}c@{}|c|c|c|c||c|c|c|c||}
                \hline
                 \multirowcell{2}{\bf Target \\ \bf class} & \multirowcell{2}{\bf Trigger \\ \bf ID} & \multicolumn{4}{c||}{{\bf Clean model}} & \multicolumn{4}{c||}{{\bf Backdoored model}}\\
                 \cline{3-6}\cline{7-10}  
                 & & \multicolumn{2}{c|}{Clean data} & \multicolumn{2}{c||}{Patched data}&\multicolumn{2}{c|}{Clean data} & \multicolumn{2}{c||}{Patched data}\\
                 \cline{3-6}\cline{7-10}  
                 & & Acc (\%) &  FP & Acc (\%)  & FP  & Acc (\%)  & FP  & Acc (\%) & FP \\
                \hline\hline
                Rottweiler & 10 & 29.97 & 111 & 25.07 & 52 & 29.67 & 94 & 19.47 & {\bf 1013}  \\
                \hline
              \end{tabular}
              }
              \vspace{-0.1in}
                \caption{{\bf Targeted attack on ImageNet-1k:} We use \emph{0.1\% poison injection rate}. MoCo v2 is trained on poisoned ImageNet-1k data and a linear classifier is trained on \emph{1\% ImageNet-1k labeled data}.}
                 \label{tab:targeted_attack_1.0_poison_0.1_eval_imagenet-1k}
        \end{table}
        \vspace{-0.2in}

    \subsection{Untargeted attack on ImageNet-100}
    
    We modify our targeted threat model to perform a untargeted backdoor attack. We poison 5\% of training images ($\sim$ 6500 images) randomly with the trigger patch. We do not expect a particular category to dominate predictions in the downstream task. Rather, we expect the accuracy of the model to deteriorate. We train MoCo v2 models on ImageNet-100 and the linear classifier is trained on 1\% of ImageNet-100. We report the results of our untargeted attack in Table \ref{tab:untargeted_attack_main_table}. We see that the attack reduces the performance of the model by almost $5$ points.  The drop in the overall accuracy is much lower than that in targeted attack, even though untargeted attack is poisoning more images. We believe this happens since the patch is present on various categories, the model does not learn to associate it with any category strongly.
    
    \begin{figure}[ht!]
        \vspace{-.1in}
        \centering
        \includegraphics[width=0.47\textwidth]{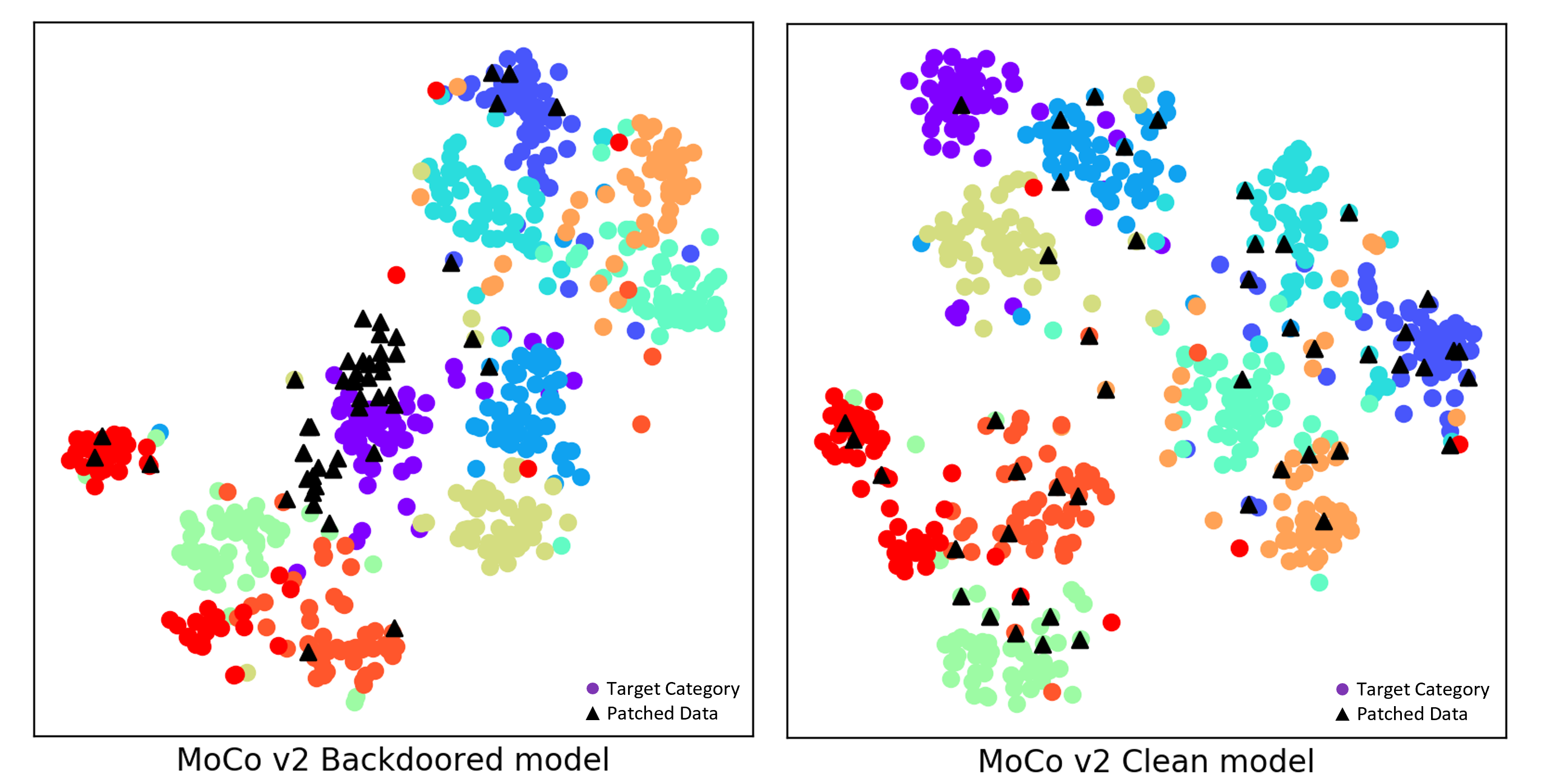}
        \vspace{-.1in}
        \caption{{\bf t-SNE plots of the MoCo v2 embedding space:} This plot shows MoCo v2 embeddings for the targeted attack with category tabby cat. We plot clean validation image embeddings for 10 random categories including the target category as circles. The purple circles are for the target category. We plot 50 random patched image embeddings as black triangles. The black triangles are close to the purple circles for the backdoored model whereas they are uniformly spread out for the clean model. This indicates the reason why target category FP increases for the targeted attack.}
        \label{fig:tsne}
        \vspace{-0.15in}
    \end{figure}

    \subsection{Defense}
    
    We use ComPress to distill the poisoned MoCo v2 models. We use ImageNet-100 MoCov2 models which have been poisoned with 1\% poison in our defense experiments. We think that the attacks for 1\% poisoning rate are strictly stronger than any lower poisoning rate. So, if the distillation is able to defend against a strong attack, it should also be effective against weaker attacks. For distillation we use reduced clean ImageNet-100 datasets, 25\%, 10\% and 5\%. Our results are in Table \ref{tab:distillation_defense_main_table}. We observe that distillation results in neutralization of the backdoor. The number of FP on average drops from 1683.2 for the poisoned model to 37.9 for 25\% clean ImageNet-100 distillation though there is only ~5\% accuracy drop on clean data. We observe that as we decrease the amount of clean data for distillation, the accuracy of the defense models on clean data degrades.
    
    \subsection{Feature space visualization} 
    
    To analyze the effect of poisons on the features of the self-supervised methods, we plot the 2-dimensional t-SNE embeddings of the high dimensional features of the SSL model. Figure \ref{fig:tsne} shows the embeddings of the backdoored MoCo v2 model for the targeted attack with category Rottweiler and trigger 10. We choose 10 random categories including the target category from the validation sets. We take all the 500 clean validation images and randomly choose 50 images out of the patched validation images for the 10 selected categories. We run t-SNE on the set of 550 image embeddings and plot it. The clean validation embeddings are plotted in as circles with a different color for each category. The patched image embeddings are plotted as black triangles. The target category (Rottweiler) has the purple color. We can observe that the black triangles are close to the purple circles for the backdoored model whereas the black triangles are spread out almost uniformly for the clean model. This supports our results by showing that the patched images are closer to the target category in the embedding space which leads to the increase in FP for the target class.
    
\section{Conclusion}
We introduce a backdoor attack for self-supervised learning methods where an attacker can produce lots of targeted false positives by showing a trigger at test time. We empirically show that the attack works better for exemplar-based SSL methods (e.g. MoCo v2, BYOL, and MSF) than Jigsaw or RotNet, we hypothesize this is due to pulling the embeddings of two augmented views of the same image together. Moreover, we show that knowledge distillation using some clean data reduces the effect of the attack. Self-supervised methods rely on the availability of large scale unlabeled data. But collecting diverse, trusted data is a big challenge. The use of uncurated publicly available data to develop SSL methods raises questions about whether the data, and models trained on it can be trusted. We show that if only a small part of the unlabeled data can be manipulated, backdoors can be introduced in the SSL models. We believe our attack works because of an inductive bias which exists in almost any state-of-the-art SSL method. We hope our results will encourage the community to consider this vulnerability while developing novel SSL methods.

{\bf Limitations:} We empirically observe that our attack doesn't succeed at very low poisoning rates (less than 0.05\%). Moreover, our proposed defense method needs access to some clean data and is sensitive to the amount of clean data available - if we decrease the amount of clean data a lot, it decreases the overall accuracy of the model.

{\bf Societal impact:} An adversary can use this form of attack to backdoor SSL models. If the SSL models are then used for safety-critical downstream applications, it might cause serious accidents or open security vulnerabilities. Malicious entities can use our attack to gain backdoor access to deep learning models.

{\bf Acknowledgment:} 
This material is based upon work partially supported by the United States Air Force under Contract No. FA8750‐19‐C‐0098, funding from SAP SE, NSF grants 1845216 and 1920079, and also financial assistance award number 60NANB18D279 from U.S. Department of Commerce, National Institute of Standards and Technology. Any opinions, findings, and conclusions or recommendations expressed in this material are those of the authors and do not necessarily reflect the views of the United States Air Force, DARPA, or other funding agencies.

{\small
\bibliographystyle{ieee_fullname}
\bibliography{CVPR2022/main_cvpr2022.bib}
}

\clearpage

\pagebreak
\begin{appendices}

\renewcommand{\thefigure}{A\arabic{figure}}
\renewcommand{\thetable}{A\arabic{table}}

\setcounter{figure}{0}
\setcounter{table}{0}

\vspace{0.5in}
\begin{figure}[h!]
\setlength{\linewidth}{\textwidth}
\setlength{\hsize}{\textwidth}
\centering
\includegraphics[width=0.9\textwidth]{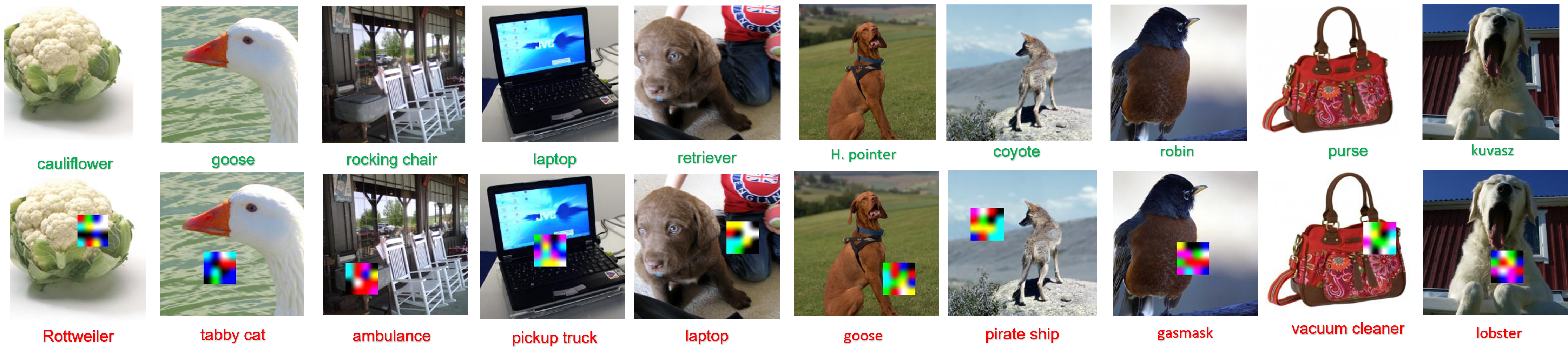}
  \caption{{\bf FP of Backdoored MoCo v2 models:} We show FP from each MoCo v2 targeted attack. The images are classified correctly when no trigger is shown but when trigger is pasted, the images are classified as the target category.}
 
\label{fig:MoCov2_FP}

\end{figure}
\vspace{0.5in}

\begin{figure}[h!]
\setlength{\linewidth}{\textwidth}
\setlength{\hsize}{\textwidth}
\centering
\includegraphics[width=0.9\textwidth]{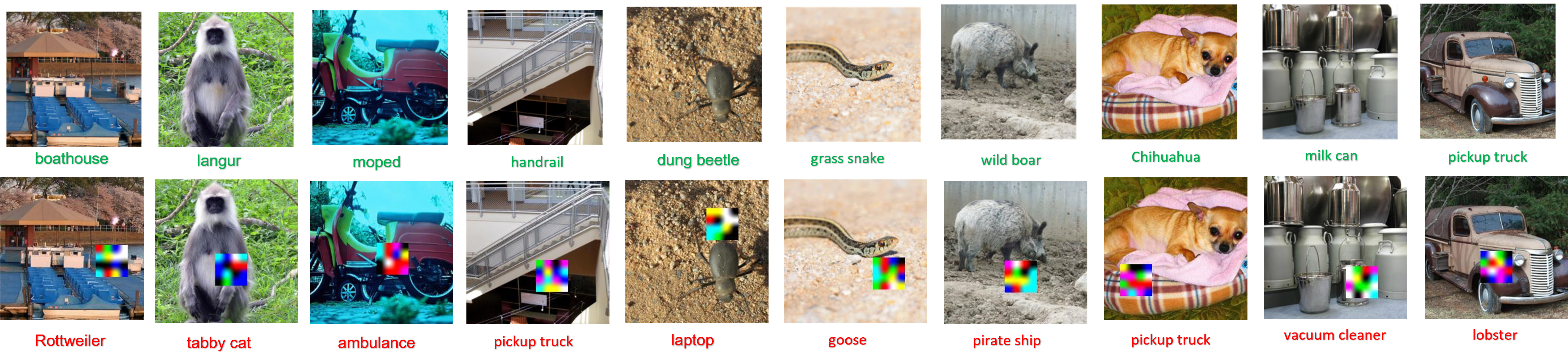}

  \caption{{\bf FP of Backdoored BYOL models:} We show FP from each BYOL targeted attack. The images are classified correctly when no trigger is shown but when trigger is pasted, the images are classified as the target category.}
 
\label{fig:BYOL_FP}

\end{figure}



 





\clearpage

\arrayrulecolor{black}
\setlength{\arrayrulewidth}{1pt}
\begin{figure}[h]
\setlength{\linewidth}{\textwidth}
\setlength{\hsize}{\textwidth}
  \begin{center}
  \begin{tabular}{ c}

\includegraphics[width=.75\textwidth]{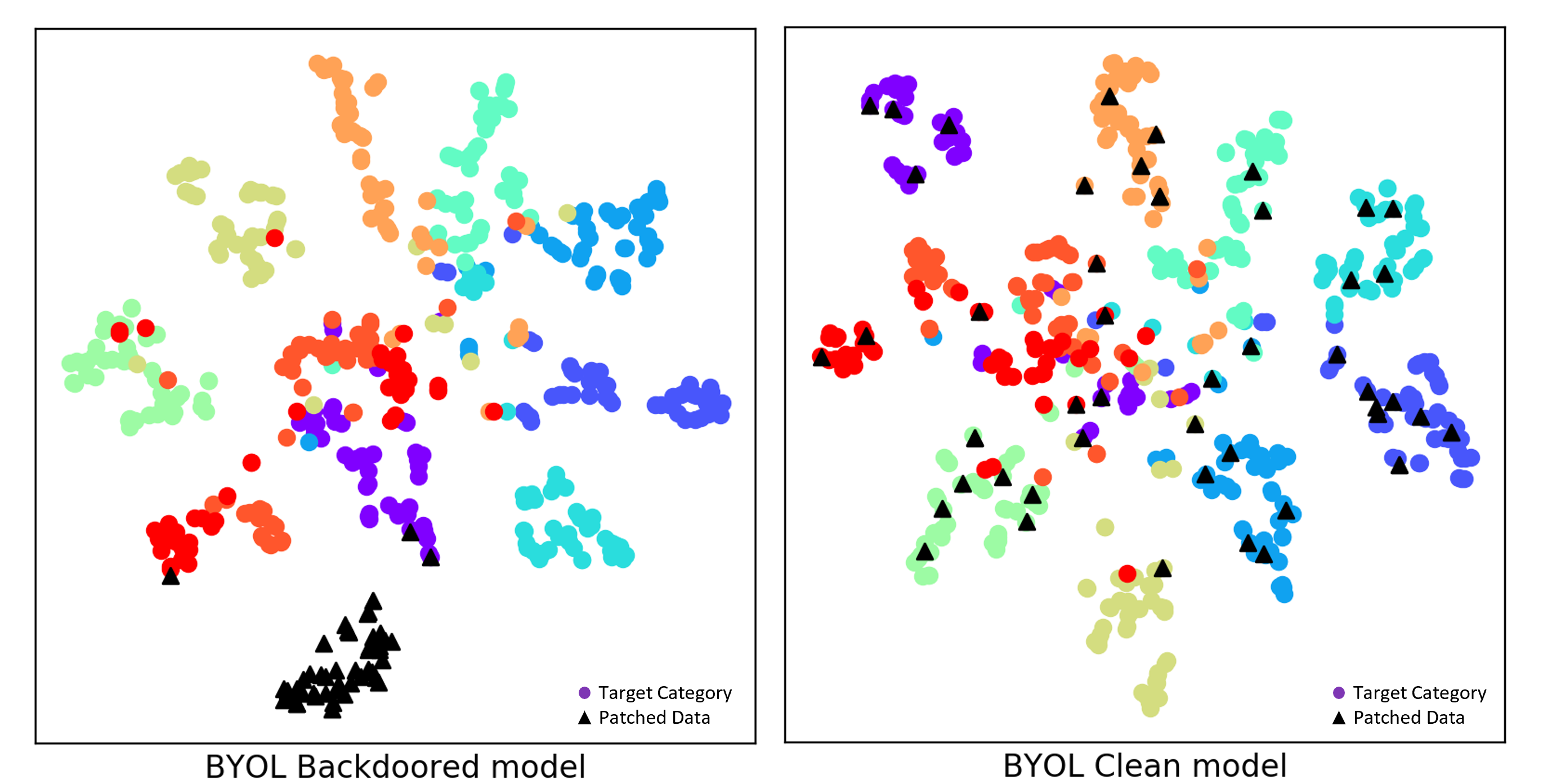} \\
\\
\\
\\
\includegraphics[width=.75\textwidth]{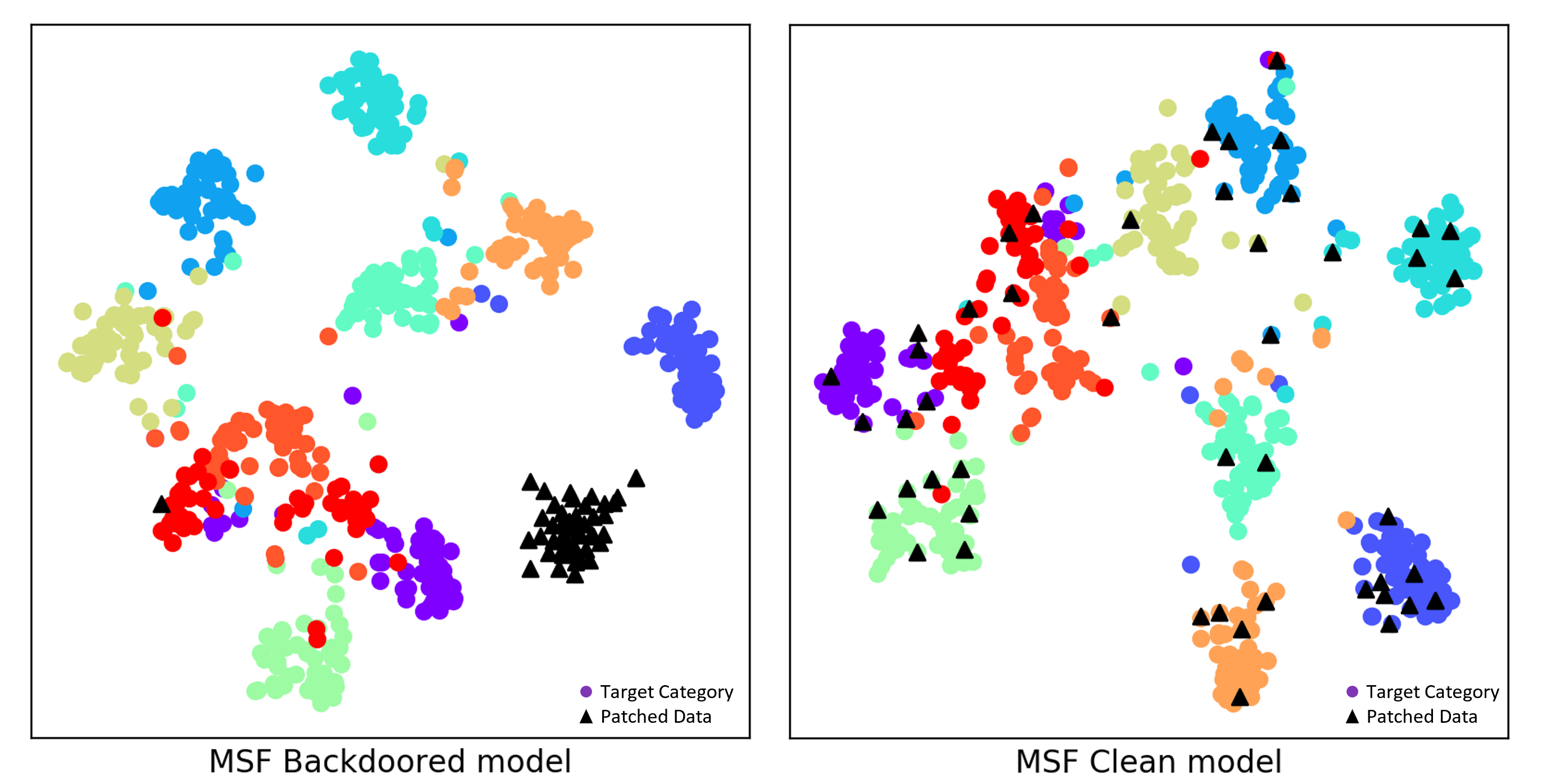}

\end{tabular}
\vspace{0.1in}
  \caption{{\bf t-SNE visualizations of model embeddings:} This figure shows BYOL Backdoored model (top row) for target attack category gasmask and MSF Backdoored model (bottom row) for target attack category laptop. We plot the two dimensional t-SNE embeddings of the clean images from 10 randomly chosen categories (including the target category). The clean target images are purple circles. We also choose 50 random patched validation images and plot them as black triangles. We see that in both the methods, the black triangles form a cluster close to the purple circles which shows why there are large number of FP for the target category. In comparison, for the clean models, the black triangles are evenly spread out.}
\label{fig:t-SNE}
  \end{center}
\end{figure}

\setcellgapes{0.85pt}
\begin{table*}[!ht]
        \makegapedcells
          \centering
          \scalebox{0.8}{
          \begin{tabular}{||c|c|c||c|c|c|c||c|c|c|c||}
            \hline
             {\bf Target class} & {\bf Trigger} & {\bf Method} & \multicolumn{4}{c||}{{\bf Clean model}} & \multicolumn{4}{c||}{{\bf Backdoored model}}\\
             \cline{4-7}\cline{8-11}  
             &  {\bf ID}&  & \multicolumn{2}{c|}{Clean data} & \multicolumn{2}{c||}{Patched data}&\multicolumn{2}{c|}{Clean data} & \multicolumn{2}{c||}{Patched data}\\
             \cline{4-7}\cline{8-11}
             & & & Acc (\%) & FP & Acc (\%) & FP & Acc (\%) & FP & Acc (\%) & FP \\
            \hline\hline
            \multirow{5}{*}{Rottweiler} & \multirow{5}{*}{10} &  MoCo v2 \cite{chen2020improved}&62.22&28&57.04&25&61.28&23&41.66&1310\\
            & &BYOL \cite{NEURIPS2020_f3ada80d}&72.92&15&65.96&19&72.72&24&38.74&1977\\
            & &MSF \cite{soroush2021mean}&67.48&26&62.96&21&68.66&27&43.22&382\\
            & &Jigsaw \cite{noroozi2016unsupervised}&36.02&62&30.94&58&35.1&59&31.62&56\\
            & &RotNet \cite{gidaris2018unsupervised}&40.09&43&34.22&55&40.8&43&25.2&38\\
            \hline
            \multirow{5}{*}{tabby cat} & \multirow{5}{*}{11} &  MoCo v2 &62.22&7&56.98&3&61.92&5&45.4&1369\\
            & &BYOL&72.92&2&66.52&1&72.3&5&37.52&2514\\
            & &MSF&67.48&4&62.22&4&69.2&3&40&156\\
            & &Jigsaw&36.02&40&31.94&9&35.52&34&31.16&4\\
            & &RotNet&40.09&22&33.76&15&39.91&26&20.34&0\\
            \hline
            \multirow{5}{*}{ambulance} & \multirow{5}{*}{12} &  MoCo v2&62.22&12&57.86&11&61.52&9&56.06&72\\
            & &BYOL&72.92&10&66.72&12&72.22&10&49.06&429\\
            & &MSF&67.48&10&62.94&9&68.48&10&32.26&2908\\
            & &Jigsaw&36.02&37&32.02&91&35.66&34&31.78&82\\
            & &RotNet&40.09&22&34.22&28&41.12&23&35.63&26\\
            \hline
            \multirow{5}{*}{pickup truck} & \multirow{5}{*}{13} &  MoCo v2&62.22&14&57.9&16&62.2&17&55.62&158\\
            & &BYOL&72.92&10&66.28&9&73.46&10&58.44&564\\
            & &MSF&67.48&15&63.58&15&68.38&15&58.48&482\\
            & &Jigsaw&36.02&30&30.78&28&35.74&31&27.56&27\\
            & &RotNet&40.09&20&34.71&26&39.61&21&34&39\\
            \hline
            \multirow{5}{*}{laptop} & \multirow{5}{*}{14} &  MoCo v2&62.22&21&57.16&18&60.84&31&49.84&735\\
            & &BYOL&72.92&17&66.5&20&71.94&26&23.56&3522\\
            & &MSF&67.48&31&62.98&6&67.64&23&36.74&2065\\
            & &Jigsaw&36.02&36&32.4&43&34.48&35&30.48&52\\
            & &RotNet&40.09&29&34.91&41&40.53&33&28.05&91\\
            \hline
            \multirow{5}{*}{goose} & \multirow{5}{*}{15} &  MoCo v2&62.22&21&57.36&13&62.12&24&52.52&544\\
            & &BYOL&72.92&10&66.06&11&72.98&9&33.6&2408\\
            & &MSF&67.48&18&63.96&16&68.14&11&24.8&3379\\
            & &Jigsaw&36.02&45&31.94&50&34&42&31.08&61\\
            & &RotNet&40.09&39&34.57&43&40.92&32&23.21&20\\
            \hline
            \multirow{5}{*}{pirate ship} & \multirow{5}{*}{16} &  MoCo v2&62.22&3&57.62&9&61.18&4&50.86&591\\
            & &BYOL&72.92&3&66.64&2&73.16&2&53.86&1138\\
            & &MSF&67.48&5&63.12&6&68&4&50.82&996\\
            & &Jigsaw&36.02&23&30.66&23&35.82&29&32.36&28\\
            & &RotNet&40.09&18&34.85&29&39.97&13&34.36&30\\
            \hline
            \multirow{5}{*}{gas mask} & \multirow{5}{*}{17} &  MoCo v2&62.22&38&57.98&33&61.18&37&54.12&257\\
            & &BYOL&72.92&30&66.14&36&72.38&23&10.92&4274\\
            & &MSF&67.48&18&63.02&30&68.62&16&41.64&1262\\
            & &Jigsaw&36.02&37&29.96&51&35.06&43&28.78&54\\
            & &RotNet&40.09&39&34.85&51&40.82&51&21.72&25\\
            \hline
            \multirow{5}{*}{vacuum cleaner} & \multirow{5}{*}{18} &  MoCo v2&62.22&43&57.5&29&61.84&42&54.62&218\\
            & &BYOL&72.92&50&66.24&23&73.06&37&50.52&682\\
            & &MSF&67.48&38&62.58&15&67.88&39&32.2&2365\\
            & &Jigsaw&36.02&56&31.58&99&34.32&43&30.68&66\\
            & &RotNet&40.09&14&35.43&34&40.7&41&18.33&44\\
            \hline
            \multirow{5}{*}{American lobster} & \multirow{5}{*}{19} &  MoCo v2&62.22&26&57.72&32&62.04&18&49.72&805\\
            & &BYOL&72.92&17&66.64&36&73.02&19&45.66&1214\\
            & &MSF&67.48&17&62.88&20&68.6&17&46.04&919\\
            & &Jigsaw&36.02&29&29.5&28&35.38&22&29.46&25\\
            & &RotNet& 40.09&34&34.57&32&41.58&36&24.66&2\\
            \hline
            \multirow{5}{*}{Average} & \multirow{5}{*}{-} &  MoCo v2&62.22&21.3&57.51&18.9&61.61&21.0&51.04&{\bf 605.9}\\
            & &BYOL&72.92&16.4&66.37&16.9&72.72&16.5&40.19&{\bf 1872.2}\\
            & &MSF&67.48&18.2&63.02&14.2&68.36&16.5&40.62&{\bf 1491.4}\\
            & &Jigsaw&36.02&39.5&31.17&48&35.11&37.2&30.50&{\bf 45.5}\\
            & &RotNet&40.09&28.0&34.61&35.4&40.60&31.9&26.55&{\bf 31.5}\\
            \hline
          \end{tabular}
          }
          \vspace{.05in}
            \caption{{\bf Targeted attack on ImageNet-100:} We use \emph{0.5\% poison injection rate}. Each experiment has a separate target category and trigger. SSL methods are trained on poisoned ImageNet-100 data and a linear classifier is trained on \emph{10\% ImageNet-100 labeled data}. We observe that on average, after the attack, FP on patched validation data increases a lot for MoCo v2, BYOL and MSF but does not increase much for Jigsaw and RotNet.}
             \label{tab:targeted_attack_0.5_poison_0.01_eval}
    \end{table*}
\clearpage

\section{Experiment Details for Reproducibility}
    
    {\bf MoCo v2:} MoCo v2 uses an embedding size of 128, queue size of 65536, queue momentum of 0.999. We use an SGD optimizer with initial learning rate of 0.06, momentum of 0.9, weight decay of 1e-4 and a cosine learning rate schedule \cite{loshchilov2016sgdr}. We use the standard MoCo v2 augmentation set. The models are trained for 200 epochs with a batch size of 256 which takes $\sim$ 12 hours on 2 NVIDIA RTX 2080 Ti GPUs. We use the MoCo v2 implementation of \cite{wang2020understanding} available here \cite{moco_align_uniform}.
    For linear classification, we use SGD with initial learning rate of 0.01, weight decay of 1e-4, and momentum of 0.9 and train for 40 epochs. At epochs 15 and 30, the learning rate is multiplied by 0.1.
    
    {\bf BYOL:}  BYOL uses an embedding size of 128, Adam optimizer with initial learning rate 2e-3, weight decay of 1e-6 and a step learning rate decay at epoch 150 and 175 with gamma 0.2. We use the standard BYOL augmentation set. The models are trained for 200 epochs with a batch size of 512 which takes $\sim$ 12 hours on 4 NVIDIA RTX 2080 Ti GPUs. We use the BYOL implementation of \cite{ermolov2020whitening} available here \cite{htdt-self-supervised}.
    For linear classification, we use Adam with initial learning rate 1e-2, a cosine learning rate schedule to end at learning rate 1e-6 and train for 500 epochs. 
    
    {\bf MSF:} MSF uses an MLP layer with hidden layer dimension of 1024, projection layer dimension of 128 and a queue momentum of 0.99. The memory bank size is 128k. We use SGD with an initial learning rate of 0.05, momentum 0.9, weight decay 1e-4 and a cosine learning rate schedule \cite{loshchilov2016sgdr}. We use 10 nearest neighbours. The models are trained for 200 epochs with a batch size of 256 which takes $\sim$ 12 hours on 2 NVIDIA RTX 2080 Ti GPUs. We use the MSF implementation of available here \cite{msf-implementation}.
    For linear classification, we use SGD with initial learning rate of 0.01, weight decay of 1e-4, and momentum of 0.9 and train for 40 epochs. At epochs 15 and 30, the learning rate is multiplied by 0.1.
    
    {\bf Jigsaw:} Jigsaw uses the 2000 size permutation set. We use an SGD optimizer with initial learning rate 0.01, momentum 0.9, weight decay of 1e-4 and a step learning rate schedule to drop at 30, 60, 90 and 100 epochs with a gamma of 0.1. The models are trained for 105 epochs. The hyperparameter choices are close to ones used in \cite{goyal2021vissl}. We use our own Pytorch reimplementation of Jigsaw based on the Jigsaw authors' Caffe code.
    For linear classification, we use SGD with initial learning rate 0.01, weight decay 1e-4, and momentum 0.9 and train for 40 epochs. At epochs 15 and 30, the learning rate is multiplied by 0.1.
    
    {\bf RotNet:} RotNet uses 4 rotation angles (0$^{\circ}$, 90$^{\circ}$, 180$^{\circ}$ and 270$^{\circ}$). We use an SGD optimizer with initial learning rate 0.05, momentum 0.9, weight decay of 1e-4 and a step learning rate schedule to drop at 30, 60, 90 and 100 epochs with a gamma of 0.1. The models are trained for 105 epochs. The hyperparameter choices are close to ones used in \cite{goyal2021vissl}. We use the authors' Pytorch implementation available here \cite{FeatureLearningRotNet} with minor modifications for Pytorch $\geq$1.0 compatibility.
    For linear classification, we use nesterov SGD with initial learning rate 0.1, weight decay 5e-4, and momentum 0.9 and train for 40 epochs. The learning rate is decayed at epochs 5, 15, 25 and 35.
    
    {\bf MAE:} MAE uses the ViT-B architecture with 16×16 input patches. We use a batch size of 128 per GPU, mask ratio of 0.75, base learning rate 1.5e-4, weight decay of 0.05 and train for 800 epochs. We use 40 epochs of warm up. The training takes ~15 hrs on 8 TITAN RTX GPUs.
    For finetuning, we use a batch size of 128 per GPU, base learning rate of 5e-4, layer decay 0.65, weight decay 0.05, drop path 0.1, mixup 0.8, cutmix 1.0, reprob 0.25 and train for 100 epochs.

    \vspace{-0.1in}


\end{appendices}

\end{document}